\documentclass{article}
\usepackage{spconf,amsmath,graphicx}
\usepackage{amsfonts}
\usepackage[dvipsnames]{xcolor}
\usepackage{booktabs}
\usepackage{caption}
\usepackage{subcaption}
\usepackage[colorlinks=true,pagebackref=true]{hyperref}
\usepackage{cleveref}
\hypersetup{
    citecolor=OliveGreen,
    linkcolor=MidnightBlue,
    urlcolor=BrickRed,
}

\newcommand{\layer}{\ell}
\newcommand{\layers}{L}
\newcommand{\InHeight}{H_\text{in}^\layer}
\newcommand{\InWidth}{W_\text{in}^\layer}
\newcommand{\InChannels}{C_\text{in}^\layer}
\newcommand{\OutHeight}{H_\text{out}^\layer}
\newcommand{\OutWidth}{W_\text{out}^\layer}
\newcommand{\OutChannels}{C_\text{out}^\layer}
\newcommand{\width}{w}
\newcommand{\height}{h}
\newcommand{\channelfirst}{i}
\newcommand{\channelsec}{j}
\newcommand{\topoloss}{\mathcal{L}_{\text{topo}}}
\newcommand{\classifloss}{\mathcal{L}_{\text{classif}}}
\newcommand{\numaxis}{N}

\title{Introducing topography in convolutional neural networks}
\name{Maxime Poli$^{\star}$ \qquad Emmanuel Dupoux$^{\star, \dagger}$ \qquad Rachid Riad$^{\star}$ \thanks{This work was performed using HPC resources from GENCI-IDRIS (Grant 2022-AD011013620). This work was supported in part by the Agence Nationale pour la Recherche (ANR-17-EURE-0017 Frontcog, ANR-10-IDEX-0001-02 PSL*, ANR19-P3IA-0001 PRAIRIE 3IA Institute) and grants from CIFAR (Learning in Machines and Brains) and Meta AI Research (Research Grant).}}

\address{$^{\star}$ENS, PSL Research University, EHESS, CNRS, INRIA \\
$^{\dagger}$Meta AI Research }

\begin{document}
%\ninept
%
\maketitle
\begin{abstract}
Parts of the brain that carry sensory tasks are organized topographically: nearby neurons are responsive to the same properties of input signals. Thus, in this work, inspired by the neuroscience literature, we proposed a new topographic inductive bias in Convolutional Neural Networks (CNNs). To achieve this, we introduced a new topographic loss and an efficient implementation to topographically organize each convolutional layer of any CNN. 
We benchmarked our new method on 4 datasets and 3 models in vision and audio tasks and showed equivalent performance to all benchmarks. Besides, we also showcased the generalizability of our topographic loss with how it can be used with different topographic organizations in CNNs. 
Finally, we demonstrated that adding the topographic inductive bias made CNNs more resistant to pruning. Our approach provides a new avenue to obtain models that are more memory efficient while maintaining better accuracy.
\end{abstract}
\begin{keywords}
Convolutional neural networks, topography, inductive bias, pruning
\end{keywords}
\vspace{-0.5em}
\section{Introduction}
\label{sec:intro}

Convolutional neural networks (CNNs) reached state-of-the-art performances in a number machine learning tasks dealing with naturalistic signals, such as image classification \cite{he2016deep,huang2017densely}, audio pattern recognition \cite{kong2020pann}, or speech recognition \cite{sercu2016very}.  

CNNs combine a number of basic operations (convolutions, non-linearity, and downsampling) to build useful hierarchical representations, and discard useless information. The introduction of these computations was in some way inspired by the brain, and in return these models can yield insights for neuroscience \cite{hassabis2017neuroscience}. Indeed, in the earlier stages of CNNs, learned filters only detect colors, edges, or orientations; whereas higher level and specialized representations are extracted later in the model \cite{zeiler2014visualizing}. This is \textit{reminiscent} of what is observed experimentally in the visual cortex \cite{felleman1991distributed}. 

However, there is no organization within a certain layer and no notion of proximity between each filter, which differs from spatial structures observed in sensory streams in the brain. Indeed, neurons in the visual \cite{zeff2007retinotopic,knapen2021topographic} and in the audio pathways \cite{barton2012orthogonal,hullett2016human} are organized in "maps", i.e. nearby neurons have similar responses. In brains, these topographic organizations of responses are observed both for (1) low-level computation such as the frequency tuning in the temporal lobe of the auditory cortex (tonotopy) \cite{khalighinejad2021functional}, but also for (2) higher-level representations such as those observed for face identification \cite{henriksson2015faciotopy} (faciotopy). Therefore, in this study, we ask the following question: Is the topographic organization of computations a useful inductive bias for CNNs?

\begin{figure*}[!ht]
    \centering
    \includegraphics[width=\linewidth]{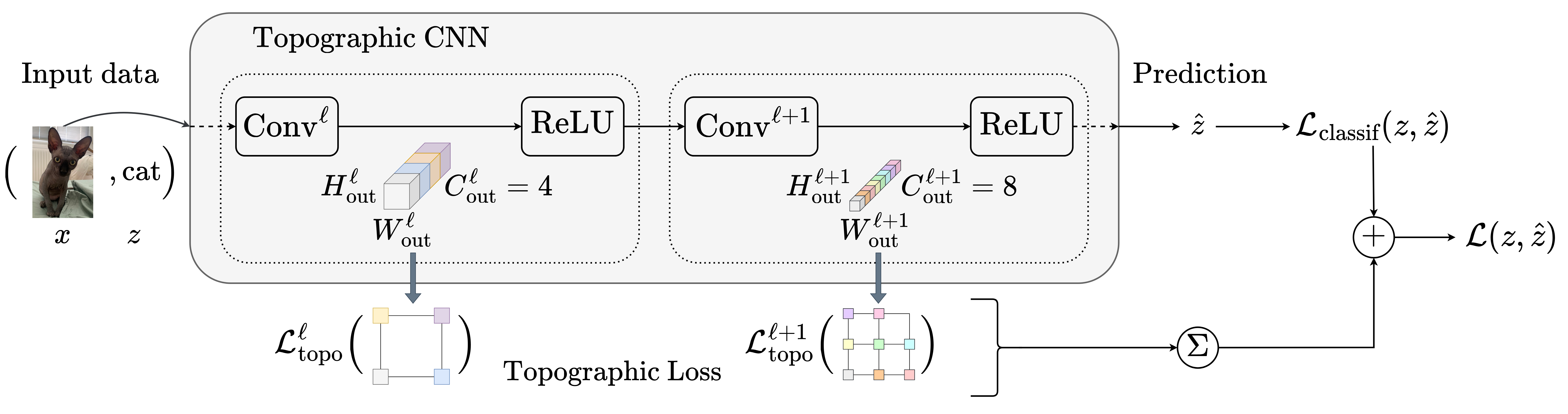}
    \caption{Our approach to introduce topography in convolutional layers with the topographic loss.}
    \label{fig:method}
\end{figure*}

In machine learning, topography as an inductive bias has been tested only in later stages of computations. It was introduced in generative models \cite{welling2002learning}, and in variational autoencoders \cite{keller2021topographic}, as a prior over the latent variables, but was only used in these higher representations and only in vision.
In CNNs, Lee et al. induced topography thanks to an auxiliary loss in the last stages of their Topographic Deep Artificial Neural Network \cite{lee2020topographic}. They mapped the positions of biological neurons in the brain to the fully connected layers of an AlexNet \cite{krizhevsky2012imagenet}. This loss is derived by considering the pairwise similarities of the activations of neurons as a function of their distance. A limitation of this study is that it relies on invasive brain recordings to assign positions to artificial neurons.

Thus, in this study, we propose a new methodology that introduces topography in convolutional layers of CNNs and which do not rely on any external data. We enforced topography with an auxiliary loss that can be used with any topographic organization. We evaluated our approach on four classification tasks, both on images and audio with three different architectures. 

Finally, the over-parametrization of CNNs is often considered as a core ingredient of their predictive power \cite{kawaguchi2017generalization}.
Yet, from an \textit{engineering} point of view, the search of memory efficient models is crucial for their deployment. On the other hand, biological neural networks need to balance between robustness and accuracy with a limited number of connections. This phenomenon, referred as \textit{synaptic pruning} in neuroscience, promotes specialized and efficient pathways. Indeed, the number of connections between biological neurons (axons and dendrites) decreases greatly between early childhood and puberty \cite{huttenlocher1979synaptic}.
From these two perspectives, we also evaluated our topographic models for their resistance to magnitude-based pruning \cite{han2015learning}, and found out that they were more robust than baselines. 

The source code to reproduce our methods and experiments is available at \href{https://github.com/bootphon/topography}{github.com/bootphon/topography}.

\vspace{-0.5em}
\section{Methods}
\label{sec:methods}
\vspace{-0.5em}
\subsection{Notations}
We consider a CNN with $\layers$ convolutional layers. The input of the network is $x$, and each convolutional layer $\layer$ receives a tensor of intermediate representations of shape $(\InChannels, \InHeight, \InWidth)$ and outputs a tensor of representations $y^{\layer}$ with a shape $(\OutChannels, \OutHeight, \OutWidth)$. $\tilde{y}^\layer$ of size $(\OutChannels, \OutHeight \OutWidth)$ is a flattened view of $y^\layer$.

\vspace{-0.5em}
\subsection{Topographic loss}
We modified existing CNNs by assigning positions to each channel of the convolutional layers. We propose that the topography in CNNs can be expressed as: nearby channels in the topography must have similar activations. In order to enforce this, we introduce an auxiliary loss, that is to be minimized in addition to the classification loss (See \cref{fig:method}).

We define a distance function $d^\layer: \{1, ..., \OutChannels\}^2 \longrightarrow \mathbb{R}^+$ where $d^\layer(i, j)$ is the euclidean distance between the positions associated to the output channels at indices $\channelfirst$ and $\channelsec$ of the layer $\layer$. We denote by $S^\layer_{\channelfirst, \channelsec}$ the cosine similarity function between those outputs channels:

\begin{equation}
    S^\layer_{\channelfirst, \channelsec}: y^\layer \mapsto \frac{1}{\|\tilde{y}^\layer_i \|_2 \|\tilde{y}^\layer_j \|_2}\sum_{\substack{1 \le \height \le \OutHeight \\ 1 \le \width \le \OutWidth}}y^\layer_{\channelfirst, \height, \width} y^\layer_{\channelsec, \height,  \width}.
\label{equ:similarity}
\end{equation}

For a given $y^\layer$, the cosine similarity matrix $S^\layer(y^\layer)$ is efficiently computed by matrix product.
All operations considered here remain differentiable and can be added to the computational graph.
The associated topographic loss $\topoloss^\layer$ at layer $\layer$, and the full loss to be minimized $\mathcal{L}$ are defined as:

\small
\setlength{\abovedisplayskip}{-5pt}
\begin{equation}
    \topoloss^\layer = \frac{2}{\OutChannels(\OutChannels-1)} \sum_{1 \le \channelfirst < \channelsec \le \OutChannels} \left( S^\layer_{\channelfirst, \channelsec} - \frac{1}{d^\layer(\channelfirst, \channelsec) +1} \right)^2 ,
\end{equation}
\normalsize

\begin{equation}
    \mathcal{L} = \classifloss + \lambda \cdot \underbrace{\frac{1}{\layers} \sum_{1 \le \layer \le \layers} \topoloss^\layer}_{\topoloss}.
\end{equation}

$\lambda$ is the weight given to the topographic loss. We chose the same target similarity profile in our loss as what was found out by \cite{lee2020topographic}. The outputs $y^\layer$ considered in the topographic loss are taken before any regularization or activation function.

\subsection{Topographic organization}
\begin{figure}[ht]
     \centering
     \begin{subfigure}[b]{0.3\columnwidth}
         \centering
         \includegraphics[width=\textwidth]{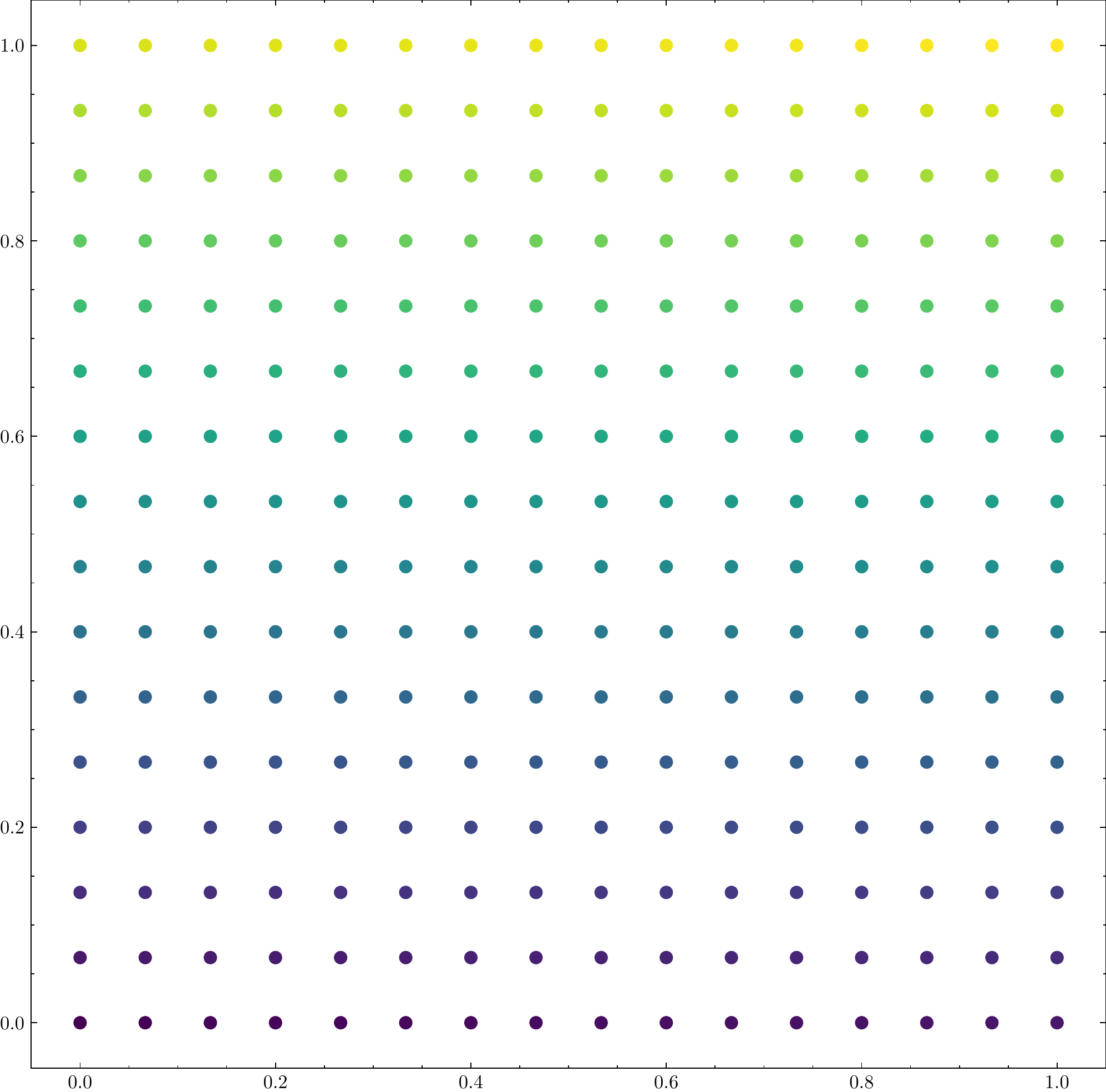}
         \caption{2D regular grid.}
         \label{fig:positions:grid2}
     \end{subfigure}
     \hfill
     \begin{subfigure}[b]{0.3\columnwidth}
         \centering
         \includegraphics[width=\textwidth]{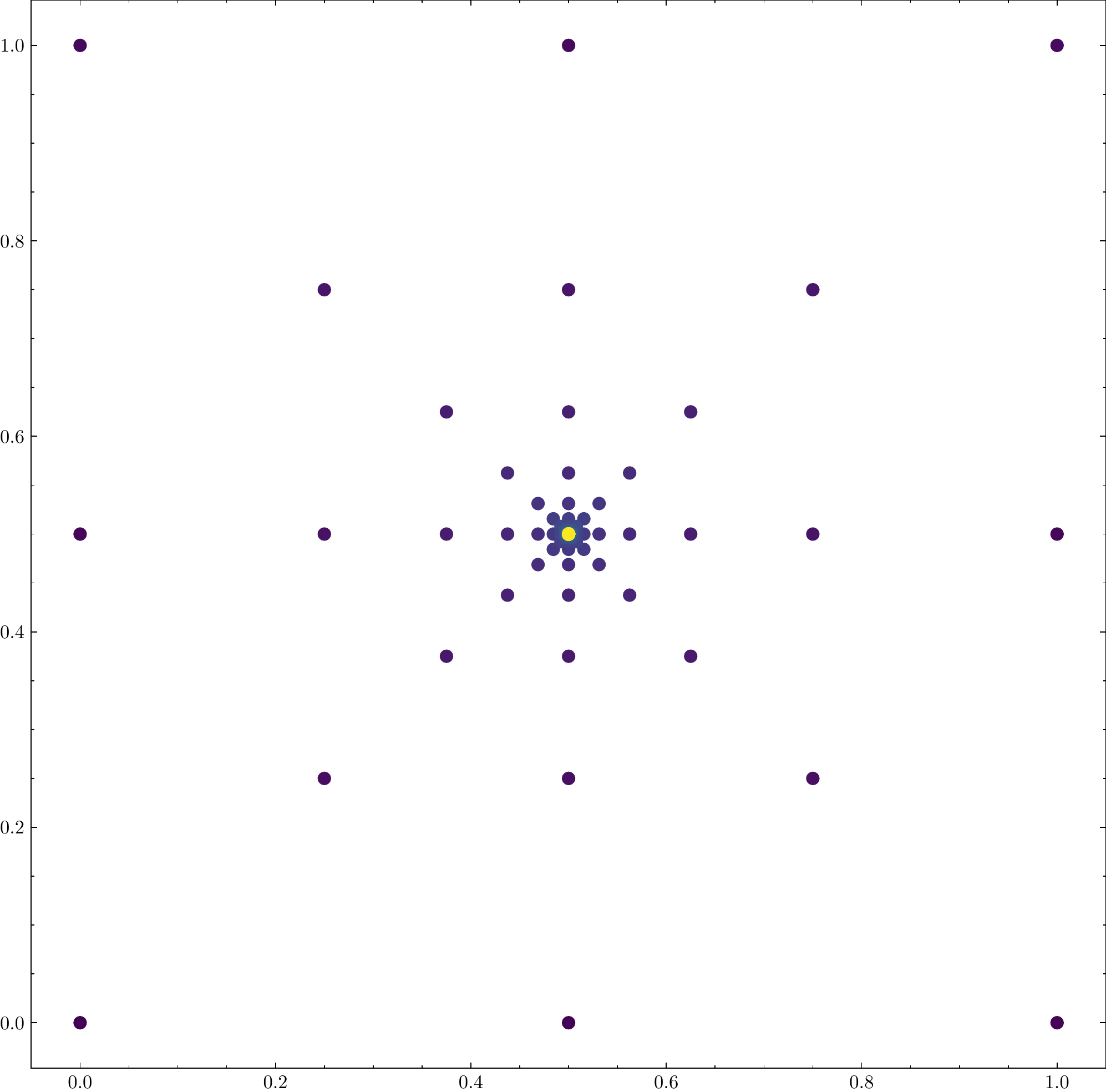}
         \caption{2D nested grid.}
         \label{fig:positions:nested2}
     \end{subfigure}
     \hfill
     \begin{subfigure}[b]{0.3\columnwidth}
         \centering
         \includegraphics[width=\textwidth]{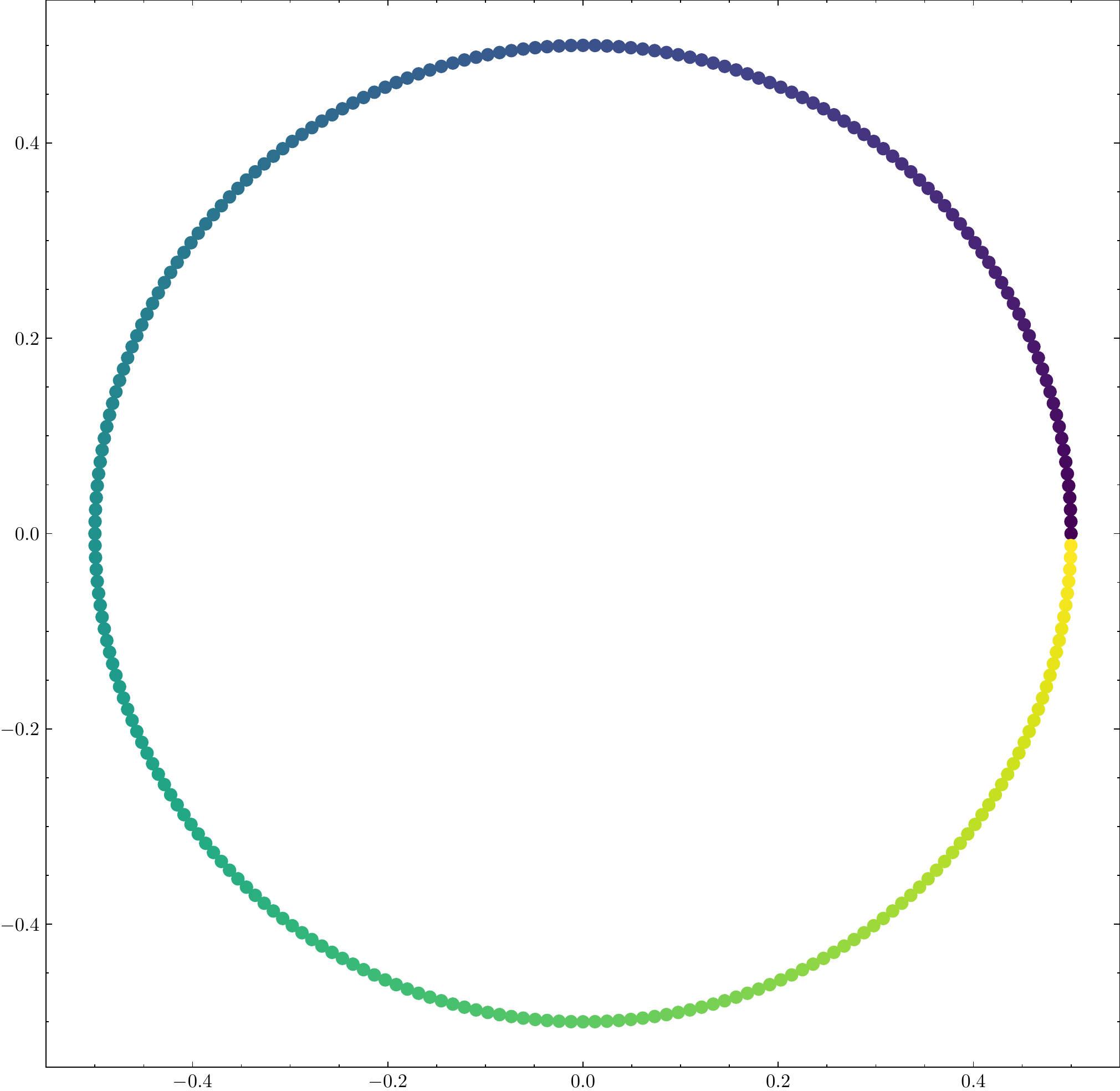}
         \caption{Circle.}
         \label{fig:positions:sphere2}
     \end{subfigure}
     \centering
     \begin{subfigure}[b]{0.3\columnwidth}
         \centering
         \includegraphics[width=\textwidth]{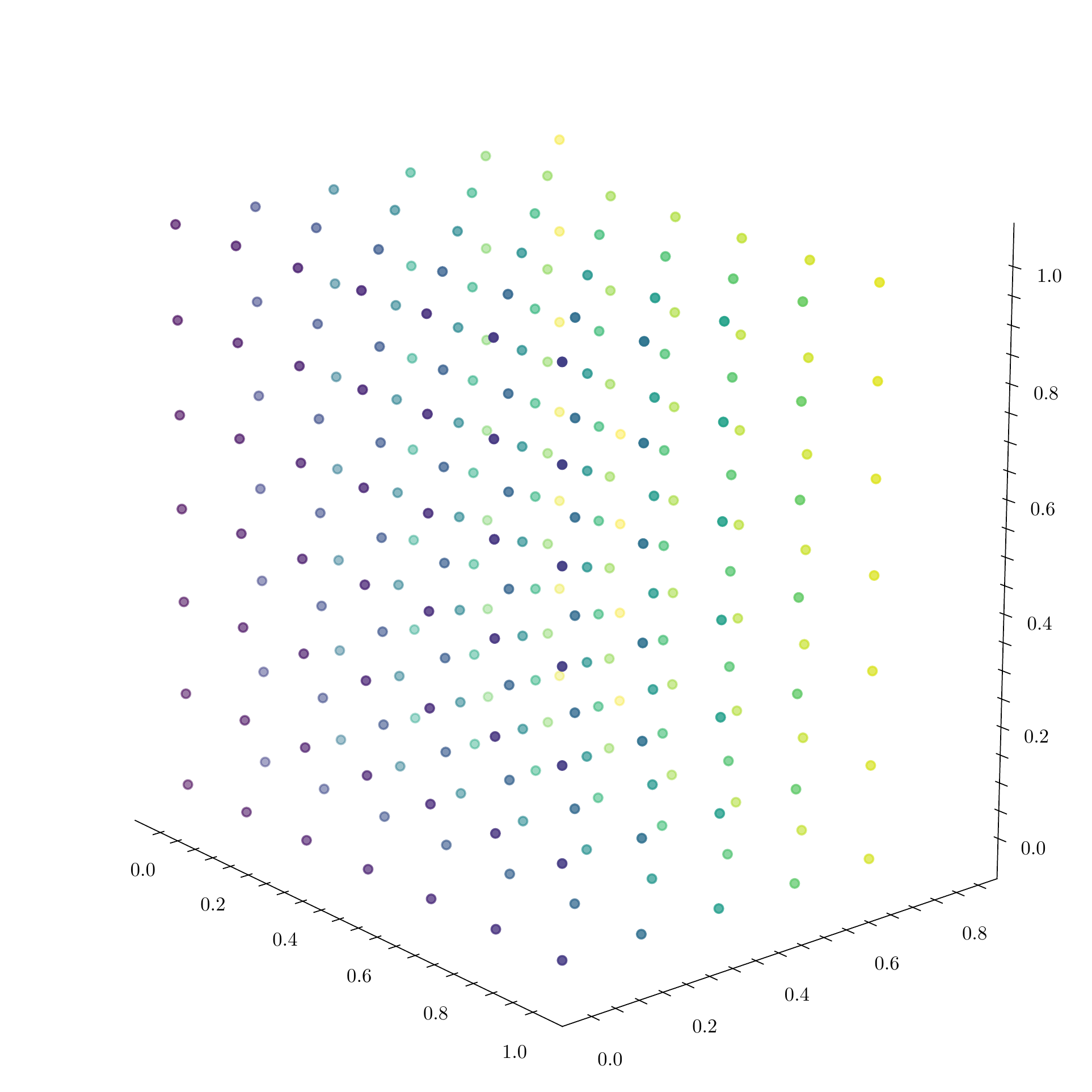}
         \caption{3D regular grid.}
         \label{fig:positions:grid3}
     \end{subfigure}
     \hfill
     \begin{subfigure}[b]{0.3\columnwidth}
         \centering
         \includegraphics[width=\textwidth]{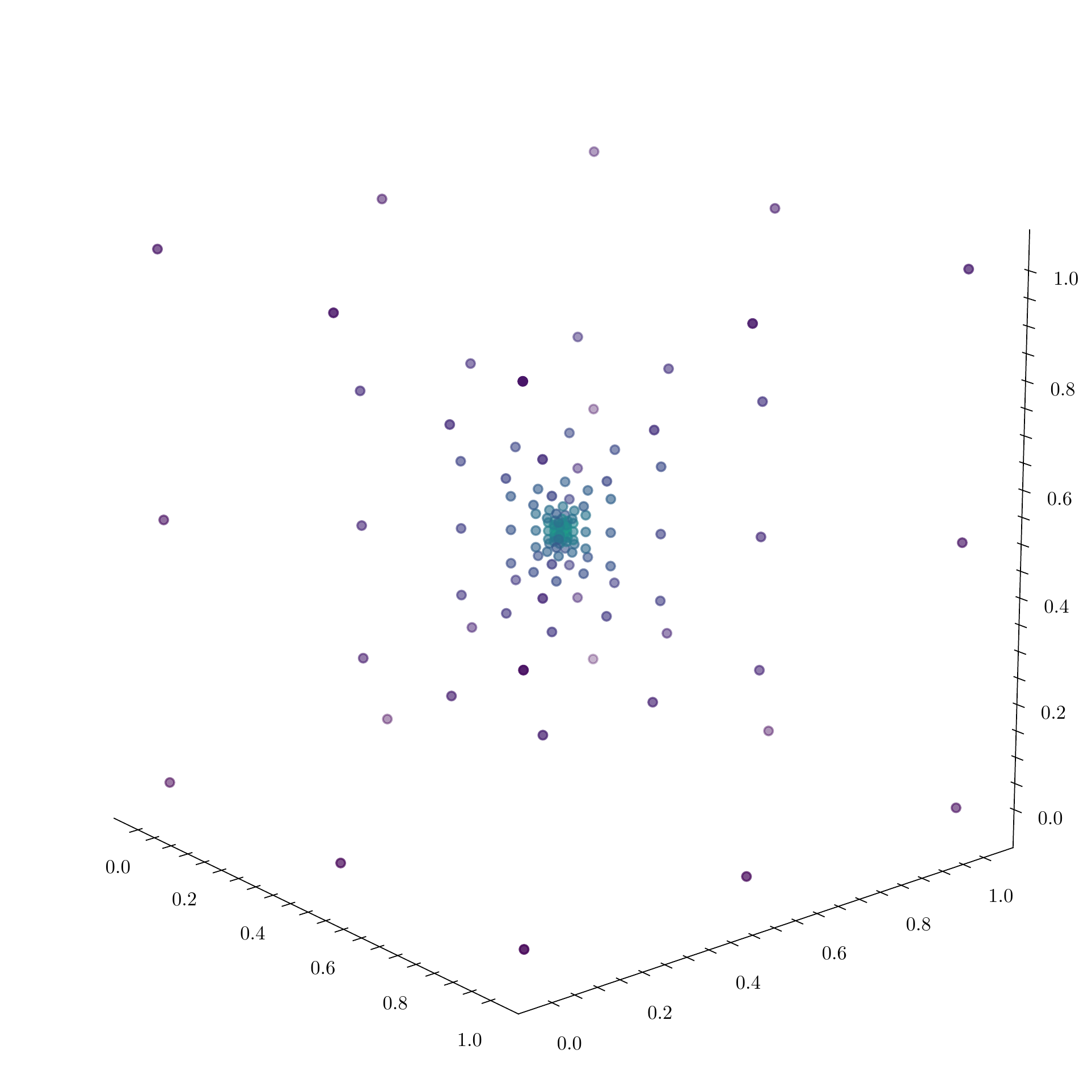}
         \caption{3D nested grid.}
         \label{fig:positions:nested3}
     \end{subfigure}
     \hfill
     \begin{subfigure}[b]{0.3\columnwidth}
         \centering
         \includegraphics[width=\textwidth]{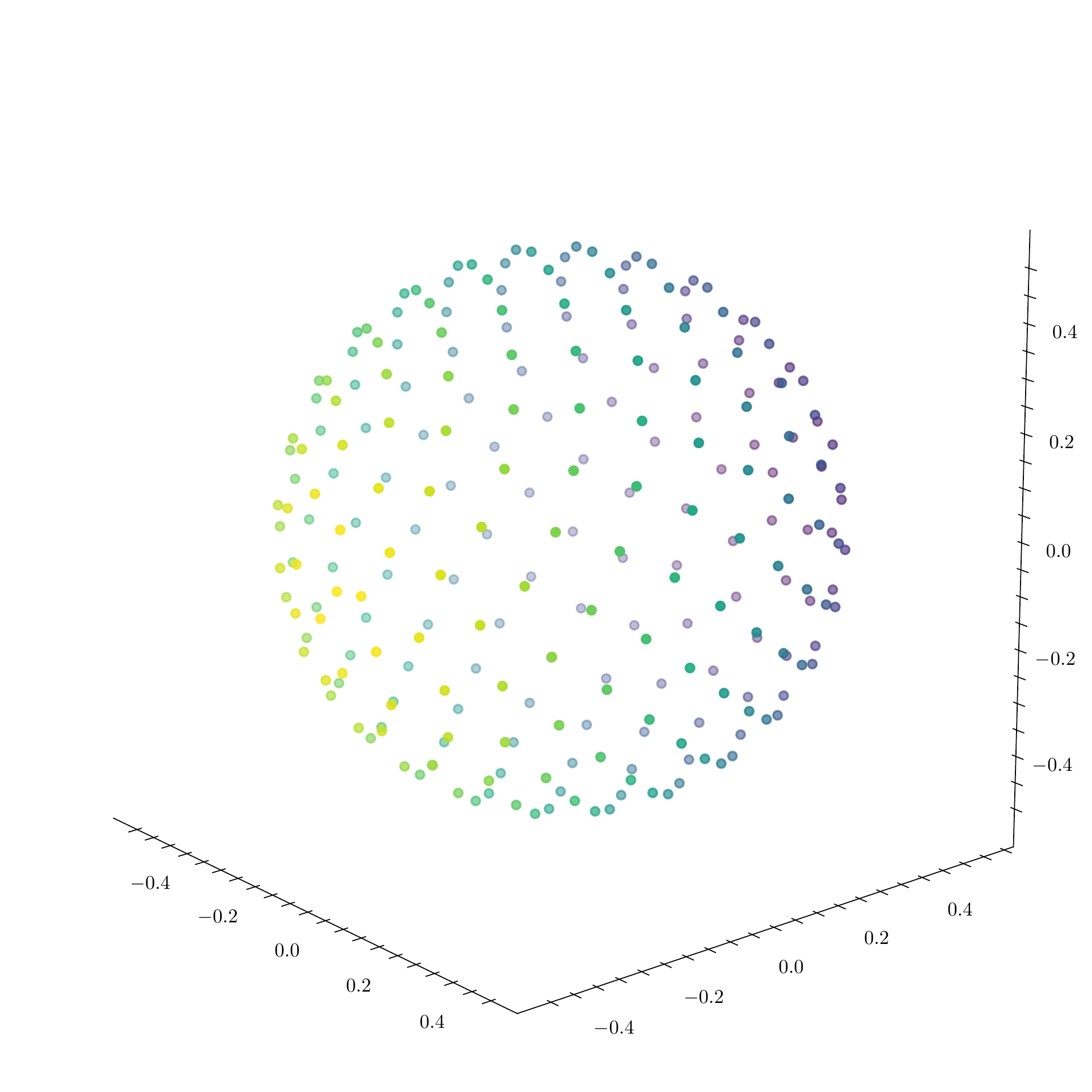}
         \caption{Sphere.}
         \label{fig:positions:sphere3}
     \end{subfigure}
        \caption{Positions of 256 channels for the different topographic organizations considered.}
        \label{fig:positions}
\end{figure}

In the formulation of the loss, the topographic organization, \textit{i.e.} the positions associated to the channels and used in $d^\layer$, was left undefined. Here, we tested and enforced different topographic organizations between the channels in convolutional layers. The schemes we used are illustrated in \cref{fig:positions}, using $256$ channels as an example.

\newpage

First, we used a 2D grid with regularly distributed positions. Coordinates are between $0$ and $1$ as the impact of this loss is weighted by the topographic weight $\lambda$ (See \cref{fig:positions:grid2}). Let $\numaxis^\layer = \lceil \sqrt{\OutChannels}\rceil$ be the number of positions per axis. For each channel $1 \le \channelfirst \le \OutChannels$, the associated position is
$(\frac{r}{\numaxis^\layer}, \frac{q}{\numaxis^\layer})$, where $q$ and $r$ are the quotient and remainder in the euclidean division of $\channelfirst-1$ by $\numaxis^\layer$.

Second, we investigated the influence of the density of positions with a variant of the regular grid: the "nested grid" (See \cref{fig:positions:nested2}). We tested such grid as density of neurons vary across and within brain regions \cite{collins2010neuron}. It is made of nested squares, with positions at each vertex and in the middle of each edge. Thus, there is a higher density in the center.

Then, as \cite{lostanlen2016deep,lostanlen2020learning} found the helix topography to be useful for audio, we also used positions taken regularly from a circle (See \cref{fig:positions:sphere2}). For each channel at index $1 \le \channelfirst \le \OutChannels$, the associated position is $\left( \cos(\frac{2 \pi}{\OutChannels} \channelfirst ), \sin (\frac{2 \pi}{\OutChannels} \channelfirst) \right)$. 
Finally, we compared these flat 2D schemes to their equivalent in 3D (See  \cref{fig:positions:grid3,fig:positions:nested3,fig:positions:sphere3}). 

\vspace{-0.5em}
\section{Experiments}
\vspace{-0.5em}
\label{sec:experiments}
We provide the full experimental setup and the procedure to reproduce our results including all hyperparameters.
\vspace{-0.5em}
\subsection{Experimental setup}
We used the ResNet-18 \cite{he2016deep}, VGG-16 \cite{simonyan2015very} and DenseNet-121 \cite{huang2017densely} architectures. Models were trained with stochastic gradient descent with a learning rate of $0.01$, a batch size of $256$, a momentum of $0.9$, and a weight decay of $0.01$. We used a learning rate scheduler with cosine annealing \cite{he2019bag} without restart and linear warmup of $30 \%$ of the number of epochs.

With the exception of the number of epochs and the data augmentation which are set according to the dataset, we kept the training procedure identical for all tasks and comparisons.
We used the introduced parameters in \cite{he2016deep} for the first layer of the ResNet-18 and DenseNet-121 with a kernel size of 3, a stride of 1, and a padding of 1. 
For ResNet-18 (resp. DenseNet) we added topography only to convolutional layers at the end of each building block (resp. dense block) to account for residual connectivity (resp. concatenation). 

For each approach, we selected the best model checkpoint and $\lambda$ based on the validation set. We report the mean accuracy and the standard deviation on the test set across 5 seeds. Our baselines on both modalities replicated \cite{he2016deep,huang2017densely,liu2015deep,huang2016deep,riad2022learning}.

\vspace{-0.5em}
\subsection{Image classification}
We benchmarked our models on CIFAR10 and CIFAR100 \cite{krizhevsky2009learning}. CIFAR10 (resp. CIFAR100) is made of $32 \times 32$ color images split into 10 (resp. 100) classes, each containing 6000 images. We used the official split with $50000$ images for the development set and $10000$ for the test set. We then split the development set into training and validation sets, with $5000$ images for the validation set. This split is the same for each run. We applied image normalization, random cropping with a padding of 4, and horizontal random flipping with a 50\% probability; and we train for 100 epochs on both datasets.

\begin{table*}[ht]
\centering
\begin{tabular}{lcccccc}
\toprule
 & \multicolumn{3}{c}{CIFAR10} & \multicolumn{3}{c}{CIFAR100} \\
 \cmidrule(lr){2-4} \cmidrule(lr){5-7}
 & DenseNet121 & ResNet18 & VGG16 & DenseNet121 & ResNet18 & VGG16 \\
\midrule
2D grid & $ \mathbf{93.58} \pm 0.21$ & $92.55 \pm 0.32$ & $93.55 \pm 0.10$ & $73.98 \pm 0.38$ & $70.86 \pm 0.31$ & $72.53 \pm 0.28$ \\
Circle & $93.46 \pm 0.39$ & $ \mathbf{92.68} \pm 0.26$ & $93.51 \pm 0.15$ & $74.05 \pm 0.26$ & $70.76 \pm 0.48$ & $72.78 \pm 0.40$ \\
2D nested & $93.42 \pm 0.18$ & $92.53 \pm 0.14$ & $93.45 \pm 0.17$ & $73.84 \pm 0.38$ & $70.93 \pm 0.19$ & $ \mathbf{72.87} \pm 0.34$ \\
3D grid & $93.47 \pm 0.18$ & $92.38 \pm 0.15$ & $ \mathbf{93.61} \pm 0.19$ & $74.02 \pm 0.46$ & $70.43 \pm 0.25$ & $72.77 \pm 0.28$ \\
Sphere & $93.55 \pm 0.19$ & $92.50 \pm 0.09$ & $93.55 \pm 0.13$ & $73.97 \pm 0.28$ & $ \mathbf{71.04} \pm 0.41$ & $72.44 \pm 0.25$ \\
3D nested & $93.55 \pm 0.17$ & $92.34 \pm 0.31$ & $93.53 \pm 0.26$ & $73.94 \pm 0.50$ & $70.91 \pm 0.26$ & $72.59 \pm 0.09$ \\
\midrule
Base & $93.43 \pm 0.13$ & $92.45 \pm 0.17$ & $93.57 \pm 0.18$ & $ \mathbf{74.10} \pm 0.22$ & $70.76 \pm 0.19$ & $72.82 \pm 0.26$ \\
\bottomrule
\end{tabular}
\caption{Mean test accuracy in $\%$ ($\pm$ std.) over 5 runs on CIFAR10 and CIFAR100. For reference, the state-of-the-art on CIFAR10 (resp. CIFAR100) is \cite{dosovitskiy2021an} (resp. \cite{foret2021sharpnessaware}) with an accuracy of $99.5\%$ (resp. $96.1\%$).}
\label{table:cifar}
\end{table*}

\vspace{-0.5em}
\subsection{Audio classification}
We evaluated our models on audio classification with two different tasks: bird sound detection (BirdDCASE) \cite{stowell2018automatic} and speech command classification (SpeechCommands) \cite{warden2018speech}. SpeechCommands is a dataset of spoken English words, while BirdDCASE is made of audio recordings with either the presence or the absence of bird sounds of any kind. BirdDCASE has the particularity that it is originally made from three development sets, each recorded under differing conditions. Those development sets have different balances of positive/negative cases, different bird species, different background sounds, and different recording equipment.

Therefore, we used the official split for SpeechCommands, whereas for BirdDCASE we split the three development datasets with the same proportion, using 80\% for the training, 10\% for the validation set, and 10\% for testing. Audio is sampled at $16$ kHz. The inputs of the networks are log mel spectrograms with a window of $25$ms computed every $10$ms, normalized by the training data statistics.

We trained for 36 epochs for BirdDCASE and 12 epochs for SpeechCommands to see 1M examples during training. On BirdDCASE, as the samples are longer than 1 second, we train with random cropped windows of 1 second. We evaluated by splitting the samples into non-overlapping 1-second windows and averaging the logits over windows, as \cite{riad2022learning}.

\begin{table*}[ht]
\centering
\begin{tabular}{lcccccc}
\toprule
 & \multicolumn{3}{c}{SpeechCommands} & \multicolumn{3}{c}{BirdDCASE} \\
 \cmidrule(lr){2-4} \cmidrule(lr){5-7}
 & DenseNet121 & ResNet18 & VGG16 & DenseNet121 & ResNet18 & VGG16 \\
\midrule
2D grid & $97.21 \pm 0.07$ & $97.35 \pm 0.04$ & $97.18 \pm 0.09$ & $82.12 \pm 0.29$ & $82.36 \pm 0.44$ & $82.51 \pm 0.22$ \\
Circle & $97.22 \pm 0.03$ & $97.35 \pm 0.11$ & $97.14 \pm 0.08$ & $82.17 \pm 0.36$ & $ \mathbf{82.44} \pm 0.28$ & $ \mathbf{82.80} \pm 0.49$ \\
2D nested & $\mathbf{97.27} \pm 0.07$ & $ \mathbf{97.38} \pm 0.09$ & $97.17 \pm 0.03$ & $82.48 \pm 0.39$ & $82.30 \pm 0.37$ & $82.68 \pm 0.34$ \\
3D grid & $ \mathbf{97.27} \pm 0.11$ & $97.33 \pm 0.15$ & $ \mathbf{97.28} \pm 0.06$ & $ \mathbf{82.58} \pm 0.08$ & $82.27 \pm 0.43$ & $82.50 \pm 0.44$ \\
Sphere & $97.19 \pm 0.13$ & $97.34 \pm 0.03$ & $97.21 \pm 0.06$ & $82.15 \pm 0.49$ & $82.40 \pm 0.22$ & $82.62 \pm 0.39$ \\
3D nested & $97.19 \pm 0.10$ & $97.33 \pm 0.07$ & $97.17 \pm 0.08$ & $82.27 \pm 0.36$ & $82.28 \pm 0.16$ & $82.67 \pm 0.49$ \\
\midrule
Base & $97.20 \pm 0.08$ & $97.36 \pm 0.07$ & $97.19 \pm 0.05$ & $82.46 \pm 0.31$ & $\mathbf{82.44} \pm 0.21$ & $82.49 \pm 0.60$ \\
\bottomrule
\end{tabular}
\caption{Mean test accuracy in $\%$ ($\pm$ std.) over 5 runs on SpeechCommands and BirdDCASE.}
\label{table:audio}
\end{table*}

\vspace{-0.5em}
\subsection{Pruning experiments}
We compared the resistance to pruning with the $L^2$ weight-based pruning \cite{han2015learning}.
For this experiment, we considered a ResNet-18 trained on CIFAR10 with a 2D grid topography and $\lambda = 1$. For each topographic layer, we pruned a fixed proportion of channels that have the weights with the lowest $L^2$-norm. We performed similarly on the corresponding layers of the baseline counterpart. We did not retrain the models after pruning.

\section{Results and discussion}

\begin{figure}[t]
    \centering
    \includegraphics[width=\linewidth]{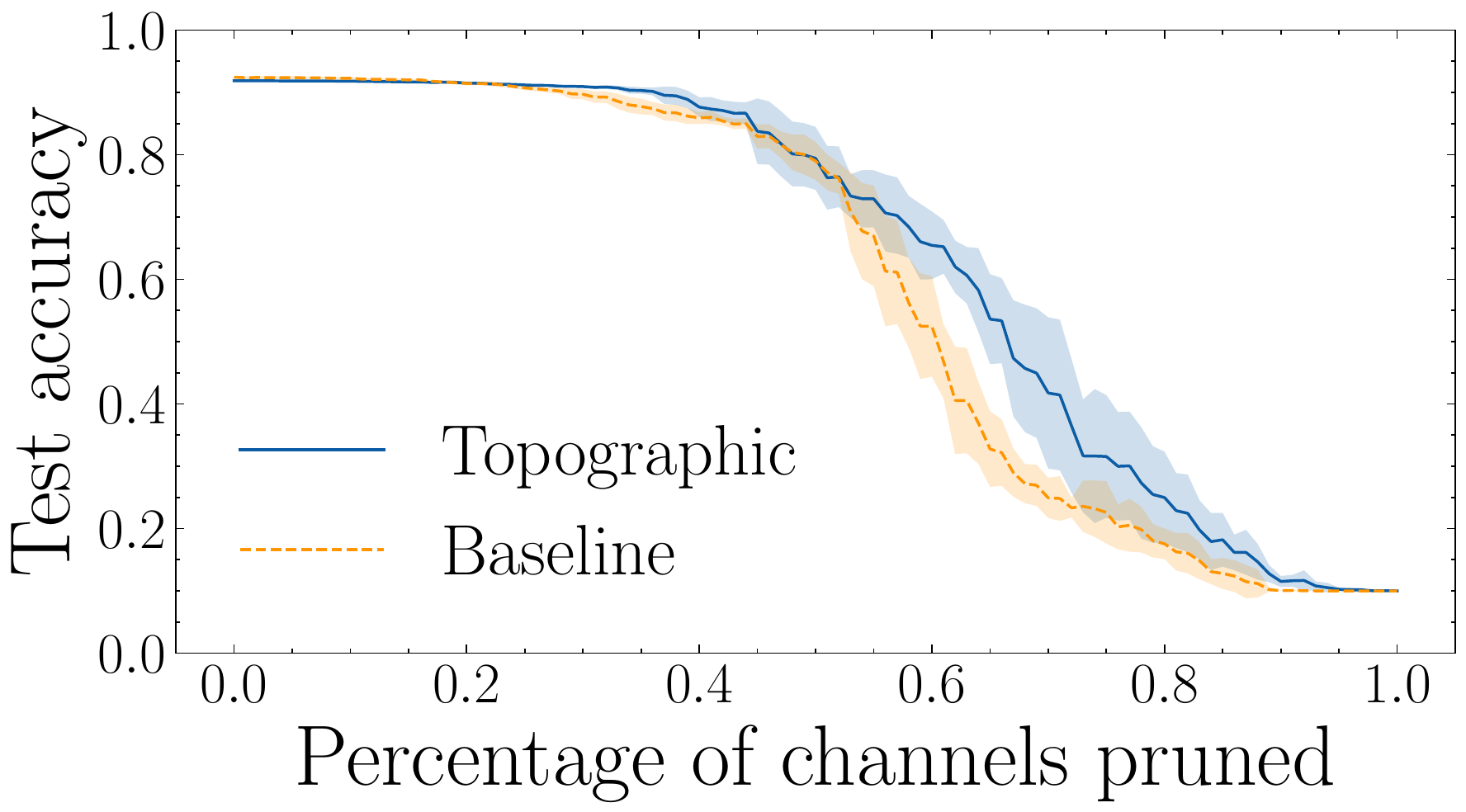}
    \caption{Mean test accuracy ($\pm$ std.) over 5 runs with $L^2$-pruning of a ResNet-18 trained on CIFAR10 with and without topography (2D grid, $\lambda=1$).}
    \vspace{-1.5em}
    \label{fig:pruning:l2}
    
\end{figure}

Introducing topography leads to results similar to baselines overall (\cref{table:cifar,table:audio}), although there are configurations with topography that provide slight improvements for the different architectures, datasets, and modalities.
In addition, for comparison, we also showed the trade-off between accuracy and proportion of pruned channels for the baseline and one topographic model (See \Cref{fig:pruning:l2}). Without pruning, the mean accuracy for this given topographic model (ResNet-18, 2D grid, $\lambda=1$) is $91.9\%$ ($92.45\%$ for the baseline). As pruning increases, our topographic model maintains better performance than the baseline. When pruning $62\%$ of channels of each topographic layer, the topographic model has $62.0\%$ test accuracy while the baseline drops to $40.6\%$. We observed similar results for the other architectures. This indicates that our topographic methods can obtain models with a better memory footprint for the same accuracy.

We did not find a better topographic organization that yields the best performance for audio or vision. This represents a potential limitation of our work as no topography stood out and is left as a modeling choice. This could be alleviated in future work if topography could be learned or selected during training. 
From a neuroscience perspective, topographic organization in the brain had been proposed as the solution to the "wiring length" minimization problem \cite{koulakov2001orientation}, mainly for biological constraints and not performance. However, our pruning results potentially suggest that topography in CNNs could also be a solution to find more efficient pathways. 

\section{Conclusion}
\label{sec:conclusion}

In this work, we introduce topography in CNNs by assigning positions to the output channels of convolutional layers and by enforcing, via an auxiliary loss, that nearby channels have similar activations. We show that, although the topographic loss is not related to the classification itself, topographic models perform similarly to the baselines. With $L^2$ weight pruning, we evidence that topography has led to a finer selection of useful channels.

In future work, instead of using predefined position schemes, we plan to also learn the topographic organization itself. Such approach could yield scientific hypotheses for a given sensory task concerning the topography itself. The learned topographies through our machine learning method could be put in parallel to the different brain regions.

\clearpage
\vfill\pagebreak
\bibliographystyle{IEEEbib}
\small{
\bibliography{reference}
}
\end{document}